\PassOptionsToPackage{unicode}{hyperref}
\PassOptionsToPackage{hyphens}{url}
\PassOptionsToPackage{dvipsnames,svgnames,x11names}{xcolor}
\documentclass[
]{article}
\usepackage{xcolor}
\usepackage{amsmath,amssymb}
\usepackage{iftex}
\usepackage[T1]{fontenc}
\usepackage[utf8]{inputenc}
\usepackage{lmodern}
\usepackage{amsmath,amssymb,amsfonts}
\usepackage{bm} 
\IfFileExists{upquote.sty}{\usepackage{upquote}}{}
\IfFileExists{microtype.sty}{
  \usepackage[]{microtype}
  \UseMicrotypeSet[protrusion]{basicmath} 
}{}
\makeatletter
\@ifundefined{KOMAClassName}{
  \IfFileExists{parskip.sty}{%
    \usepackage{parskip}
  }{
    \setlength{\parindent}{0pt}
    \setlength{\parskip}{6pt plus 2pt minus 1pt}}
}{
  \KOMAoptions{parskip=half}}
\makeatother
\makeatletter
\ifx\paragraph\undefined\else
  \let\oldparagraph\paragraph
  \renewcommand{\paragraph}{
    \@ifstar
      \xxxParagraphStar
      \xxxParagraphNoStar
  }
  \newcommand{\xxxParagraphStar}[1]{\oldparagraph*{#1}\mbox{}}
  \newcommand{\xxxParagraphNoStar}[1]{\oldparagraph{#1}\mbox{}}
\fi
\ifx\subparagraph\undefined\else
  \let\oldsubparagraph\subparagraph
  \renewcommand{\subparagraph}{
    \@ifstar
      \xxxSubParagraphStar
      \xxxSubParagraphNoStar
  }
  \newcommand{\xxxSubParagraphStar}[1]{\oldsubparagraph*{#1}\mbox{}}
  \newcommand{\xxxSubParagraphNoStar}[1]{\oldsubparagraph{#1}\mbox{}}
\fi
\makeatother

\usepackage{longtable,booktabs,array}
\usepackage{calc} 
\usepackage{etoolbox}
\makeatletter
\patchcmd\longtable{\par}{\if@noskipsec\mbox{}\fi\par}{}{}
\makeatother
\IfFileExists{footnotehyper.sty}{\usepackage{footnotehyper}}{\usepackage{footnote}}
\makesavenoteenv{longtable}
\usepackage{graphicx}
\makeatletter
\newsavebox\pandoc@box
\newcommand*\pandocbounded[1]{
  \sbox\pandoc@box{#1}%
  \Gscale@div\@tempa{\textheight}{\dimexpr\ht\pandoc@box+\dp\pandoc@box\relax}%
  \Gscale@div\@tempb{\linewidth}{\wd\pandoc@box}%
  \ifdim\@tempb\p@<\@tempa\p@\let\@tempa\@tempb\fi
  \ifdim\@tempa\p@<\p@\scalebox{\@tempa}{\usebox\pandoc@box}%
  \else\usebox{\pandoc@box}%
  \fi%
}
\def\fps@figure{htbp}
\makeatother

\NewDocumentCommand\citeproctext{}{}

\makeatletter
 \let\@cite@ofmt\@firstofone
 \def\@biblabel#1{}
 \def\@cite#1#2{{#1\if@tempswa , #2\fi}}
\makeatother
\newlength{\cslhangindent}
\newlength{\csllabelwidth}
\newenvironment{CSLReferences}[2] 
 {\begin{list}{}{%
  \setlength{\itemindent}{0pt}
  \setlength{\leftmargin}{0pt}
  \setlength{\parsep}{0pt}
  \ifodd #1
   \setlength{\leftmargin}{\cslhangindent}
   \setlength{\itemindent}{-1\cslhangindent}
  \fi
  \setlength{\itemsep}{#2\baselineskip}}}
 {\end{list}}
\usepackage{calc}

\newcommand{\CSLLeftMargin}[1]{\parbox[t]{\csllabelwidth}{\strut#1\strut}}
\newcommand{\CSLRightInline}[1]{\parbox[t]{\linewidth - \csllabelwidth}{\strut#1\strut}}

\usepackage{arxiv}
\usepackage{orcidlink}
\usepackage{amsmath}
\usepackage[T1]{fontenc}
\makeatletter
\@ifpackageloaded{caption}{}{\usepackage{caption}}
\AtBeginDocument{%
\ifdefined\contentsname
  \renewcommand*\contentsname{Table of contents}
\else
  \newcommand\contentsname{Table of contents}
\fi
\ifdefined\listfigurename
  \renewcommand*\listfigurename{List of Figures}
\else
  \newcommand\listfigurename{List of Figures}
\fi
\ifdefined\listtablename
  \renewcommand*\listtablename{List of Tables}
\else
  \newcommand\listtablename{List of Tables}
\fi
\ifdefined\figurename
  \renewcommand*\figurename{Figure}
\else
  \newcommand\figurename{Figure}
\fi
\ifdefined\tablename
  \renewcommand*\tablename{Table}
\else
  \newcommand\tablename{Table}
\fi
}
\@ifpackageloaded{float}{}{\usepackage{float}}
\floatstyle{ruled}
\@ifundefined{c@chapter}{\newfloat{codelisting}{h}{lop}}{\newfloat{codelisting}{h}{lop}[chapter]}
\floatname{codelisting}{Listing}

\makeatother
\makeatletter
\@ifpackageloaded{caption}{}{\usepackage{caption}}
\@ifpackageloaded{subcaption}{}{\usepackage{subcaption}}
\makeatother
\makeatletter
\@ifpackageloaded{algorithm}{}{\usepackage{algorithm}}
\makeatother
\makeatletter
\@ifpackageloaded{algpseudocode}{}{\usepackage{algpseudocode}}
\makeatother
\makeatletter
\@ifpackageloaded{caption}{}{\usepackage{caption}}
\makeatother
    \makeatletter
    \newcommand\fs@nocaption{
      \def\@fs@cfont{\bfseries}
      \let\@fs@capt\floatc@plain
      \def\@fs@pre{}%
      \def\@fs@post{\kern2pt\hrule}%
      \def\@fs@mid{\hrule\kern2pt}%
      \let\@fs@iftopcapt\iftrue}
    \makeatother
  
\usepackage{bookmark}
\IfFileExists{xurl.sty}{\usepackage{xurl}}{} 
\hypersetup{
  pdftitle={SAGE: Subsurface AI-driven Geostatistical Extraction with proxy posterior},
  pdfauthor={Huseyin Tuna Erdinc; Ipsita Bhar; Rafael Orozco; Thales de Sauza Oliveira; Felix J. Herrmann},
  colorlinks=true,
  linkcolor={blue},
  filecolor={Maroon},
  citecolor={Blue},
  urlcolor={Blue},
  pdfcreator={LaTeX via pandoc}}

\title{SAGE: Subsurface AI-driven Geostatistical Extraction with proxy
posterior}
\def\asep{\\\\\\ } 
\def\asep{\And }
\author{\textbf{Huseyin Tuna Erdinc}\\\\Georgia Institute of
Technology\\\\\asep\textbf{Ipsita Bhar}\\\\Georgia Institute of
Technology\\\\\asep\textbf{Rafael Orozco}\\\\Oxy\\\\\asep\textbf{Thales Souza}\\\\Huber Engineered Materials\\\\\asep\textbf{Felix
J. Herrmann}\\\\Georgia Institute of Technology\\\\}
\date{}
\begin{document}
\maketitle
\begin{abstract}
Recent advances in generative networks have enabled new approaches to
subsurface velocity model synthesis, offering a compelling alternative
to traditional methods such as Full Waveform Inversion. However, these
approaches predominantly rely on the availability of large-scale
datasets of high-quality, geologically realistic subsurface velocity
models, which are often difficult to obtain in practice. We introduce
SAGE, a novel framework for statistically consistent proxy velocity
generation from incomplete observations, specifically sparse well logs
and migrated seismic images. During training, SAGE learns a proxy
posterior over velocity models conditioned on both modalities (wells and
seismic); at inference, it produces full-resolution velocity fields
conditioned solely on migrated images, with well information implicitly
encoded in the learned distribution. This enables the generation of
geologically plausible and statistically accurate velocity realizations.
We validate SAGE on both synthetic and field datasets, demonstrating its
ability to capture complex subsurface variability under limited
observational constraints. Furthermore, samples drawn from the learned
proxy distribution can be leveraged to train downstream networks,
supporting inversion workflows. Overall, SAGE provides a scalable and
data-efficient pathway toward learning geological proxy posterior for
seismic imaging and inversion. Repo link:
https://github.com/slimgroup/SAGE.
\end{abstract}

\floatname{algorithm}{Algorithm}

\newcommand{\argmin}{\mathop{\mathrm{argmin}\,}\limits}
\newcommand{\argmax}{\mathop{\mathrm{argmax}\,}\limits}

\[
\def\textsc#1{\dosc#1\csod} 
\def\dosc#1#2\csod{{\rm #1{\small #2}}} 
\]

\section{Introduction}\label{introduction}

Characterization of subsurface physical properties (e.g., velocity,
density, compressional wavespeed, impedence, Poisson ratio) is
foundational to quantitative geoscience. These parameters govern wave
propagation and provide key inputs to applications including resource
exploration{[}1{]}, reservoir monitoring {[}2{]}, {[}3{]}, and
prediction of subsurface fluid flow and storage {[}4{]}, {[}5{]}. In
particular, within seismic exploration workflows, accurate velocity
models are indispensable for both imaging and inversion, as they
directly regulate the kinematics and recorded wavefields consequently,
the fidelity of subsurface reconstruction and property estimation.
Traditionally, velocity estimation has been addressed via Full Waveform
Inversion (FWI) {[}6{]}; however, its high-dimensional, nonlinear, and
ill-posed nature renders the problem challenging in practice.

Recent advances in generative networks and machine learning offer a
data-driven answer to this challenge by introducing training approaches
that allow creation of scalable and high-performance algorithms for
velocity model building. Various approaches have been developed that
formulate the task as Bayesian posterior inference with uncertainty
quantification {[}7{]}, {[}8{]}. Subsequent works incorporate
physics-informed representations and summary statistics, such as
reverse-time migration (RTM) {[}9{]}, {[}10{]} or common image gathers
{[}11{]}, {[}12{]}, to improve conditioning. Other methods leverage
explicitly trained networks as priors and perform inversion via
sampling-based schemes coupled with the forward operator {[}13{]},
{[}14{]}. Despite these advances, all the approaches fundamentally rely
on having access to large-scale, geologically realistic, high-resolution
velocity models for training {[}15{]}, which are often prohibitively
scarce.

Inspired by self-supervised learning literature {[}16{]} and approaches
{[}17{]}, {[}18{]}, we propose SAGE, a novel approach that circumvents
the need for fully observed velocity models. SAGE learns a proxy
posterior over velocity fields using only migrated images and occluded
velocity models (well-log measurements). The key contributions of SAGE
are: (1) learning a proxy posterior from paired incomplete velocity and
migrated image supervision; (2) amortized synthesis of high-resolution
velocity realizations conditioned solely on migrated at inference; (3)
utilization as a data sample generator for training downstream,
task-specific networks; and (4) efficient fine-tuning on field migrated
image and well-log data.

\section{Theory and Methodology}\label{theory-and-methodology}

\subsection{Seismic imaging and Bayesian
inference}\label{seismic-imaging-and-bayesian-inference}

The estimation of subsurface properties, such as the acoustic wavespeed
field \(\mathbf{x}\), can be formulated as an inverse problem from
recorded seismic data \(\mathbf{d}\). The relationship between the model
parameters and the observed data is governed by the seismic wave
equation through a nonlinear forward modeling operator:
\(\mathbf{d} =\mathcal{F}(\mathbf{x}) + \boldsymbol{\epsilon},\boldsymbol{\epsilon} \sim p(\boldsymbol{\epsilon})\)
where \(\mathcal{F}\) denotes the nonlinear forward operator mapping
subsurface parameters to predicted wavefield measurements at receiver
locations, and \(\boldsymbol{\epsilon}\) represents band-limited
acquisition noise. Recovering the subsurface model \(\mathbf{x}\) from
\(\mathbf{d}\) is challenging due to the strong nonlinearity of the
forward operator, the presence of a nontrivial null space, and the
compounding effects of noise and modeling errors. To mitigate these
difficulties, we employ summary statistics {[}19{]} that preserve key
salient structural information while transforming the data into a more
interpretable representation. In seismic imaging, a commonly used
summary statistic is the reverse-time migration (RTM) image. Let
\(\mathbf{x}_0\) denote a background model and let
\(\mathbf{J}(\mathbf{x}_0) = \left.\frac{\partial \mathcal{F}}{\partial \mathbf{x}}\right|_{\mathbf{x}_0}\)
denote the Jacobian of the forward operator evaluated at
\(\mathbf{x}_0\). The RTM image is given by:
\(\mathbf{y}=\mathbf{J}(\mathbf{x}_0)^ \top\left(\mathcal{F}(\mathbf{x}_0) - \mathbf{d}\right)\).
Despite this transformation, the inverse problem remains ill-posed:
multiple velocity model, \(\mathbf{x}\), can explain the observations
equally well. This ambiguity motivates a probabilistic formulation.
Within a Bayesian framework, the posterior distribution over model
parameters is given by by Bayes' rule:
\(p(\mathbf{x} \mid \mathbf{y}) = \frac{p(\mathbf{y} \mid \mathbf{x}) p(\mathbf{x})}{p(\mathbf{y})},\)
where \(p(\mathbf{x})\) denotes the prior distribution describing
information about the velocity model before observing the data, and
\(p(\mathbf{y} \mid \mathbf{x})\) is the likelihood, which measures how
probable the observed RTM image \(\mathbf{y}\) is for a given model
\(\mathbf{x}\). Classical Bayesian inference relies on evaluating the
likelihood and prior to generate posterior samples. In this work, we
instead estimate a proxy posterior \(\tilde{p}(\mathbf{x}|\mathbf{y})\)
under the assumption that only occluded (partially observed)
measurements of prior samples are available.

\subsection{Simulation-based inference with conditional score-based
networks}\label{simulation-based-inference-with-conditional-score-based-networks}

Simulation-based inference (SBI) is a framework that allows the training
of surrogates for posterior using neural estimators {[}20{]}. The key
idea is to use numerical simulators to generate training pairs
\(\mathcal{D} = \{ (\mathbf{x}^{(i)}, \mathbf{y}^{(i)}) \}_{i=1}^{N}\),
where each pair consists of a set of subsurface velocity
\(\mathbf{x}^{(i)}\) and the corresponding observation
\(\mathbf{y}^{(i)}\) derived using the forward simulation and linearized
imaging. The resulting pairs \((\mathbf{x}^{(i)}, \mathbf{y}^{(i)})\)
are then used to train a conditional generative network, which learns
the posterior distribution of the velocities conditioned on migrated
images. In this study, we will use conditional score-based generative
networks in an SBI setting {[}21{]}.

Score-based diffusion networks are density estimators that learn the
score function of a sequence of noise-perturbed distributions.
Specifically, the network learns
\(\nabla_{\mathbf{x}_t} \log p_t(\mathbf{x}_t)\) where
\(p_t(\mathbf{x}_t)\) denotes the distribution of the data after
corruption with Gaussian noise. In the variance-exploding formulation
used in this work, the perturbed distribution can be written as
\(p_t(\mathbf{x}_t) = p(\mathbf{x}) * \mathcal{N}(0, \sigma(t)^2 \mathbf{I})\)
where \(t \in [0,1]\) denotes diffusion time and \(\sigma(t)\) is a
monotonically increasing noise schedule. This construction corresponds
to progressively perturbing samples from the data distribution with
Gaussian noise of increasing variance, thereby inducing a continuum of
distributions that interpolate between the data (\(\mathbf{x}\)) and an
isotropic Gaussian. The forward diffusion process that generates these
perturbed samples is governed by a stochastic differential equation
{[}22{]}. This process induces a monotonic increase in variance,
progressively perturbing samples until the distribution approaches an
isotropic Gaussian. Once the score function
\(\nabla_{\mathbf{x}_t} \log p_t(\mathbf{x}_t)\) has been learned,
samples from the data distribution can be generated by simulating a
reverse-time stochastic differential equation {[}22{]}. As proposed by
{[}23{]}, we adopt a simplified score-learning objective with
\(\sigma(t) = t\). The training objective is given by: \[
\hat{\boldsymbol{\theta}} 
= \arg\min_{\boldsymbol{\theta}} 
\mathbb{E}_{\mathbf{x} \sim p(\mathbf{x})}
\mathbb{E}_{\sigma \sim \mathrm{LogNormal}(P_{\mathrm{mean}}, P_{\mathrm{std}}^2)} 
\mathbb{E}_{\mathbf{n} \sim \mathcal{N}(0, \mathbf{I})}
\left\| D_{\boldsymbol{\theta}}(\mathbf{x} + \sigma \mathbf{n}, \sigma) - \mathbf{x} \right\|_2^2,
\] where the denoising network
\(D_{\boldsymbol{\theta}}(\mathbf{x}_t, \sigma)\), with learnable
parameters \(\boldsymbol{\theta}\), is trained to recover the clean
signal \(\mathbf{x}\) from its noisy counterpart
\(\mathbf{x}_t = \mathbf{x} + \sigma \mathbf{n}\), where
\(\mathbf{n} \sim \mathcal{N}(0, \mathbf{I})\). The parameters
\(P_{\mathrm{mean}}\) and \(P_{\mathrm{std}}\) define the
\(\mathrm{LogNormal}\) distribution governing the noise levels
\(\sigma\). The score function of the perturbed distribution can then be
approximated as
\(\nabla_{\mathbf{x}_t} \log p_t(\mathbf{x}_t) \approx \frac{D_{\boldsymbol{\theta}}(\mathbf{x}_t, \sigma) - \mathbf{x}_t}{\sigma^2}\).
This formulation corresponds to the unconditional setting. To model
conditional distributions, we extend the framework to estimate the
conditional score
\(\nabla_{\mathbf{x}_t} \log p_t(\mathbf{x}_t \mid \mathbf{y})\) where
\(\mathbf{y}\) denotes the conditioning variable (e.g., migrated
images). The corresponding training objective is modified to incorporate
conditioning as follows: \[
\hat{\boldsymbol{\theta}} = \arg\min_{\boldsymbol{\theta}} \mathbb{E}_{(\mathbf{x},\mathbf{y},\sigma,\mathbf{n}) \sim p(\mathbf{x},\mathbf{y},\sigma,\mathbf{n})} \left\| D_{\boldsymbol{\theta}}(\mathbf{x} + \sigma \mathbf{n}, \mathbf{y}, \sigma) - \mathbf{x} \right\|_2^2.
\] However, SBI approaches for posterior estimation typically require
access to samples from the prior distribution \(p(\mathbf{x})\). In the
proposed SAGE framework, we relax this requirement by learning from
partially observed realizations of \(\mathbf{x}\) (velocities).

\subsection{SAGE Framework}\label{sage-framework}

\begin{figure*}[!t]
\centering
\begin{tabular}{@{}cccccc@{}}
\includegraphics[width=0.16\textwidth]{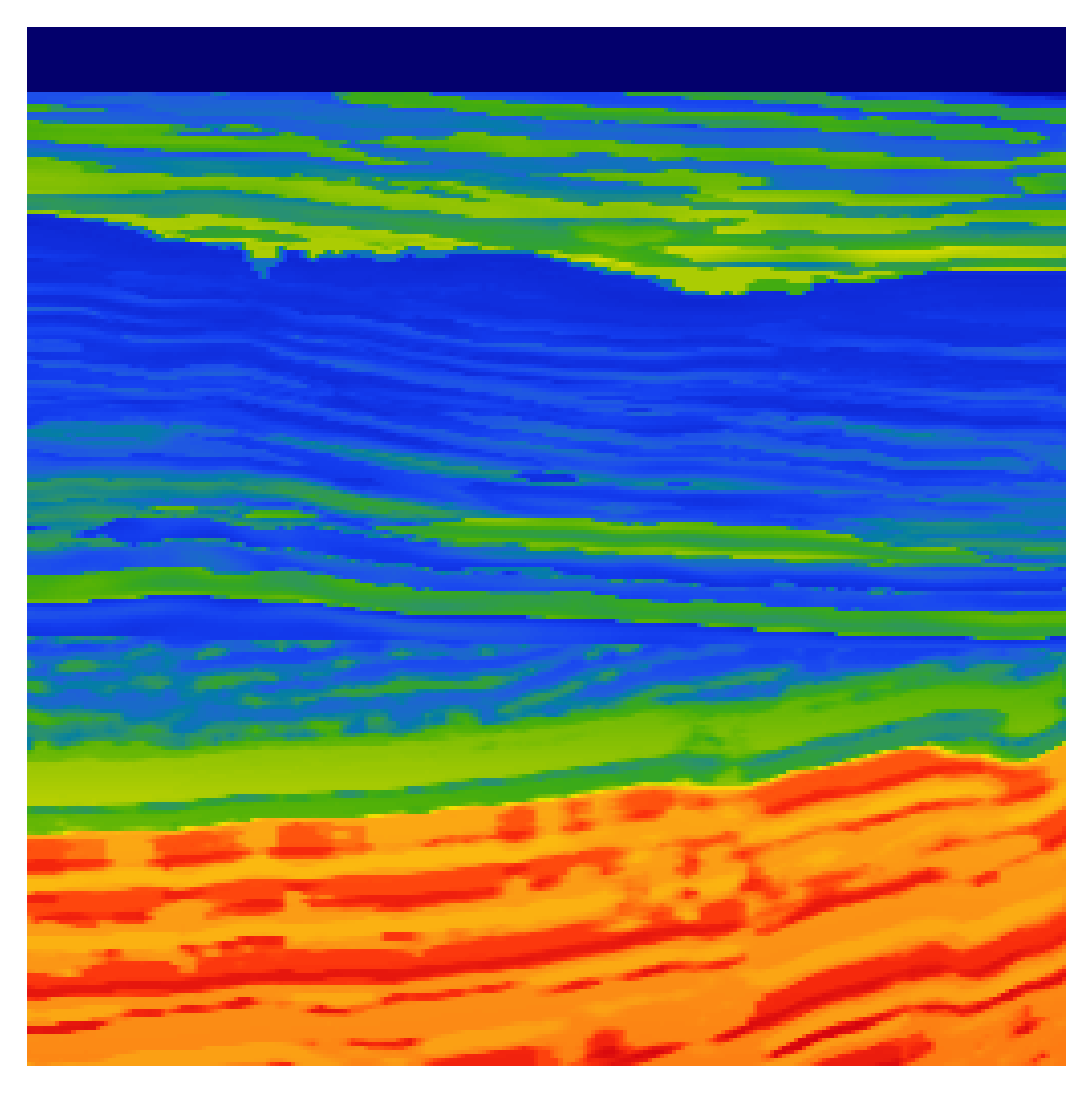} &
\includegraphics[width=0.16\textwidth]{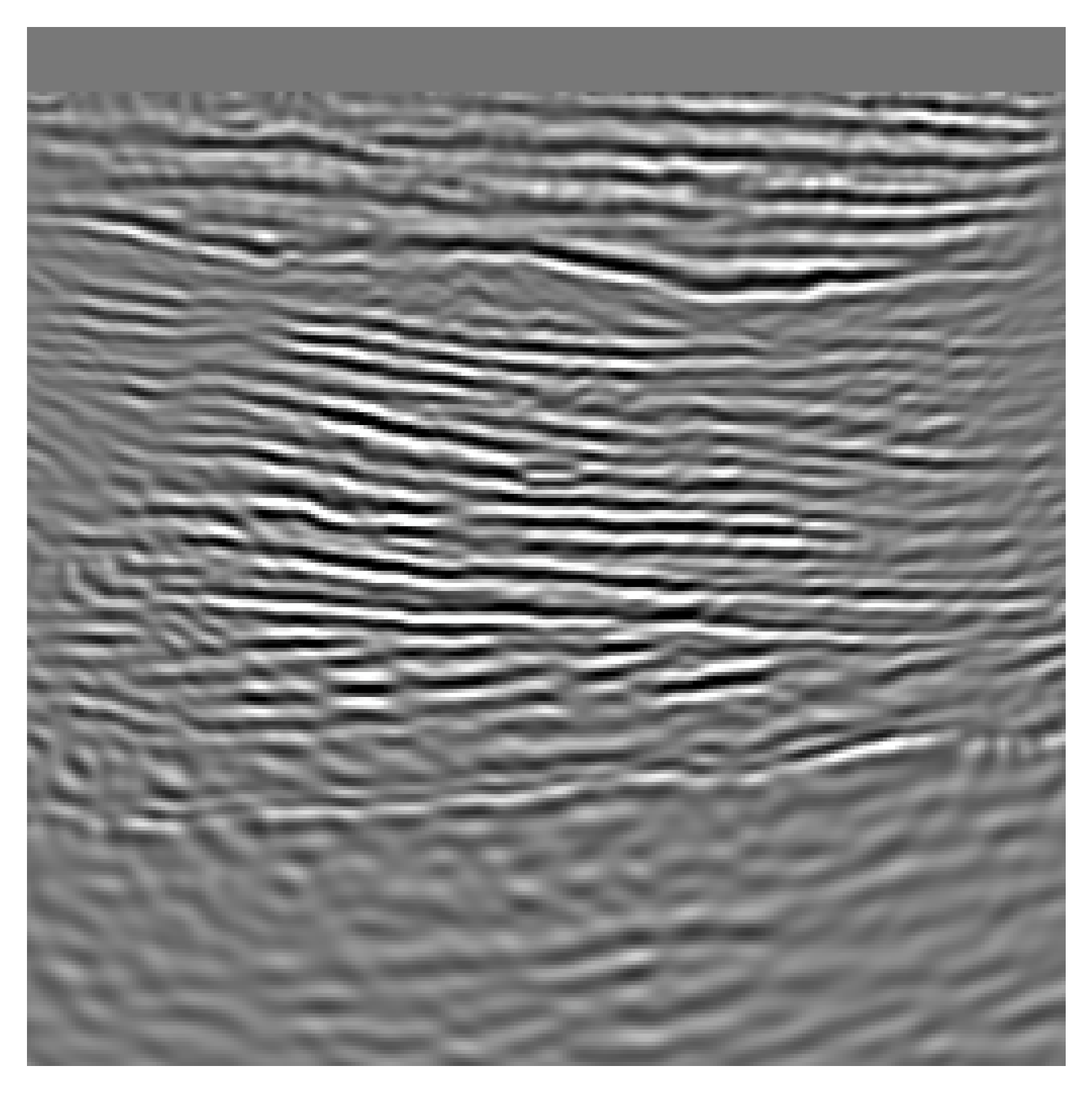} &
\includegraphics[width=0.16\textwidth]{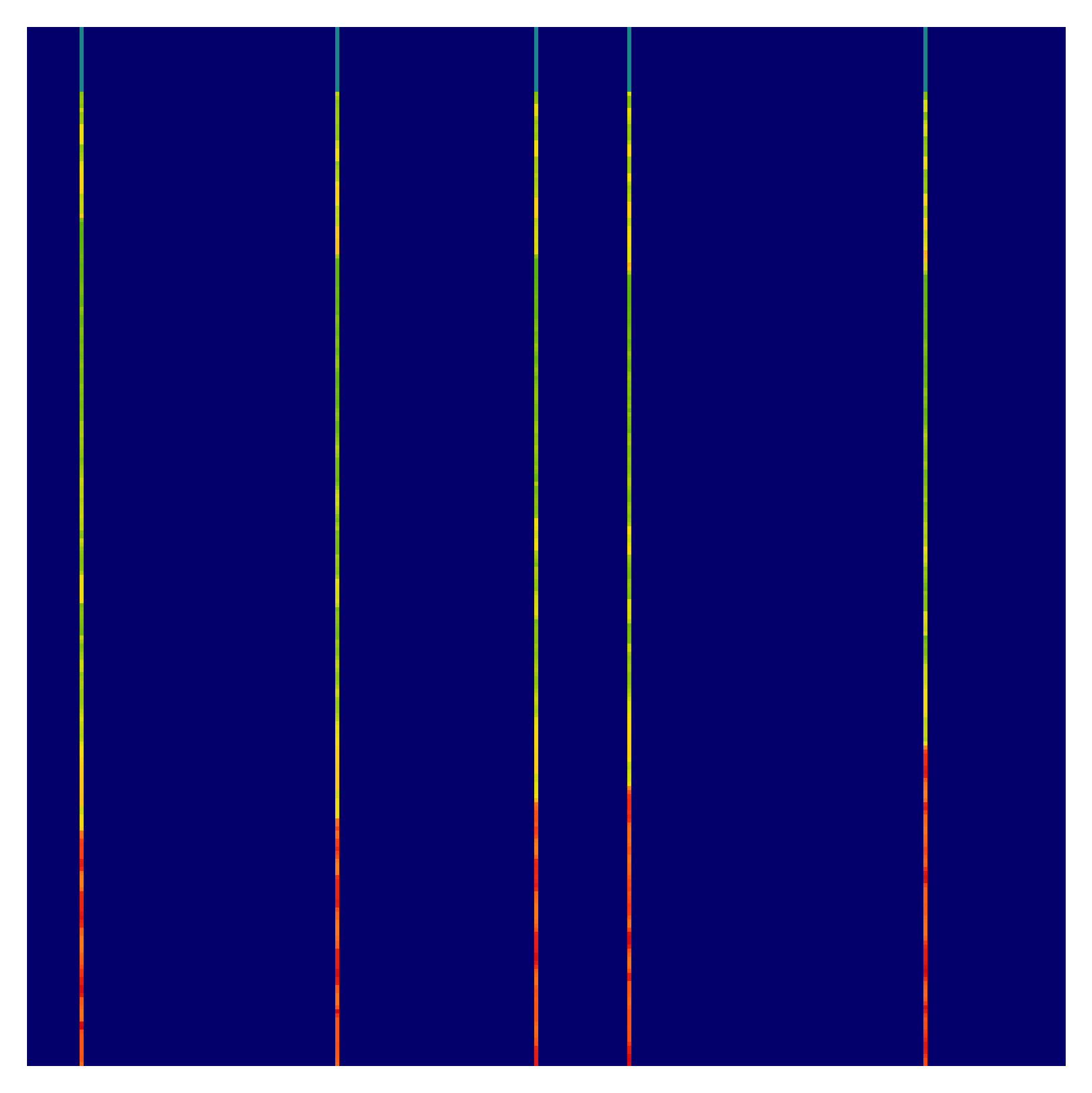} &
\includegraphics[width=0.16\textwidth]{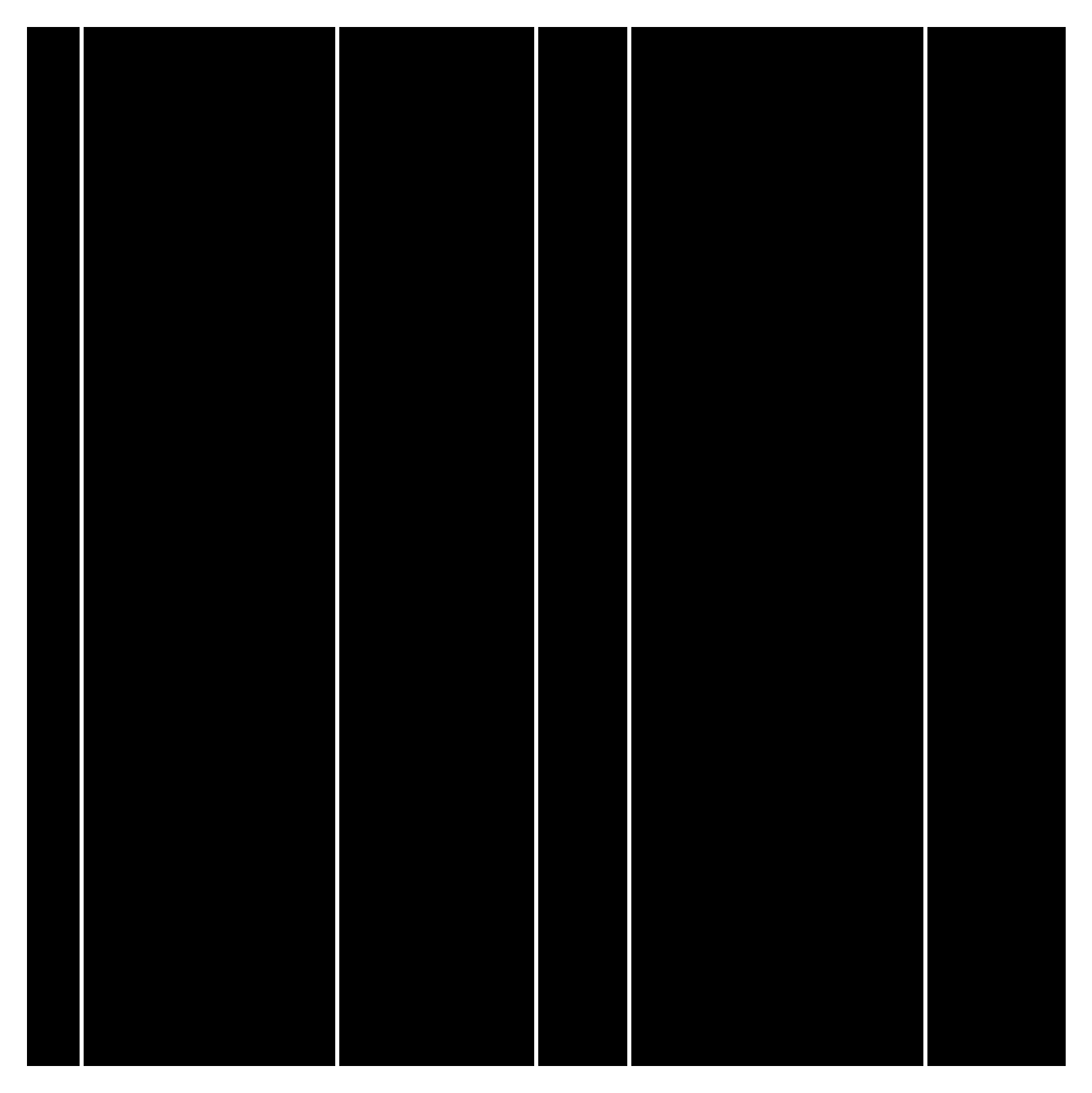} &
\includegraphics[width=0.16\textwidth]{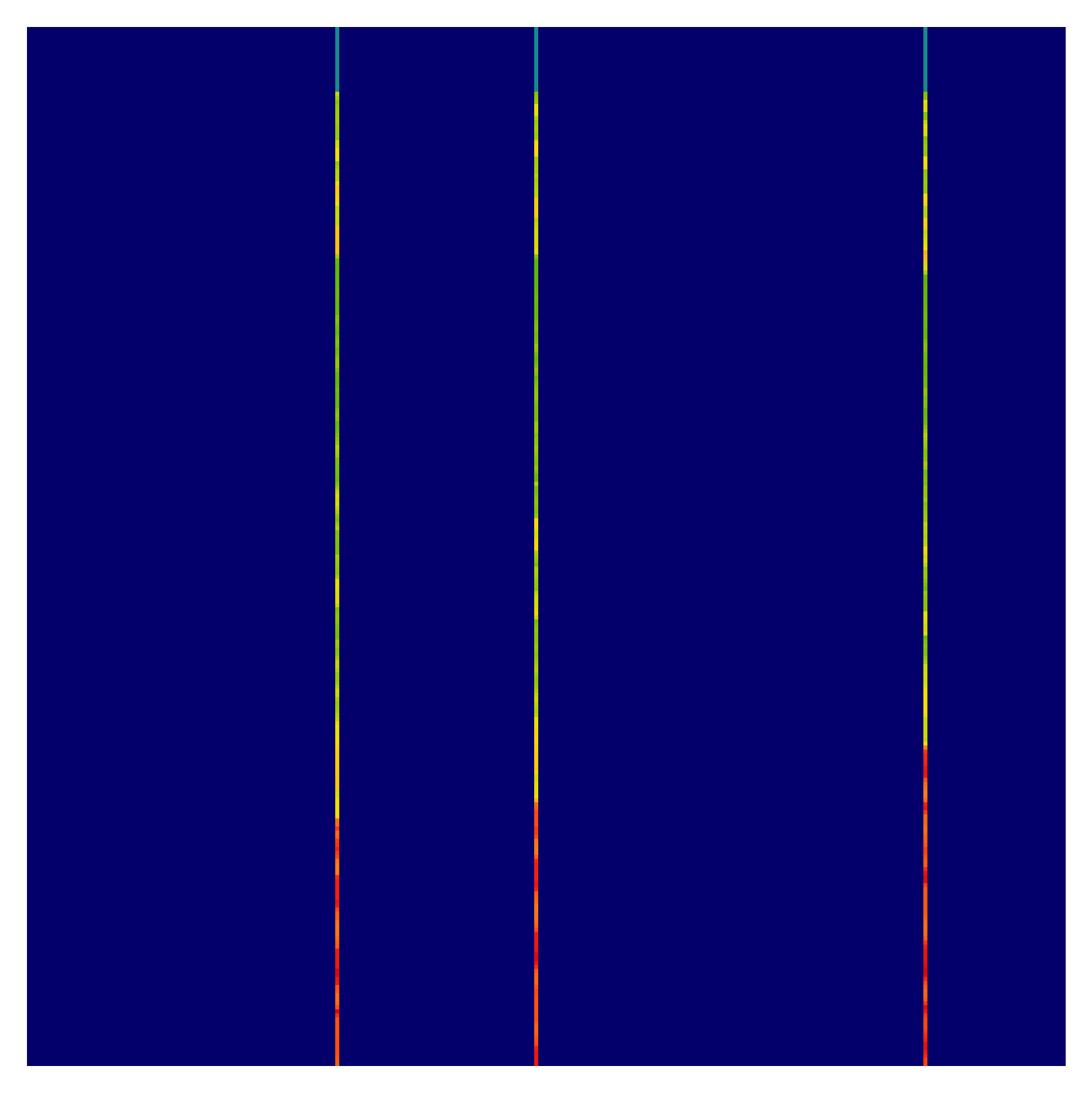} &
\includegraphics[width=0.16\textwidth]{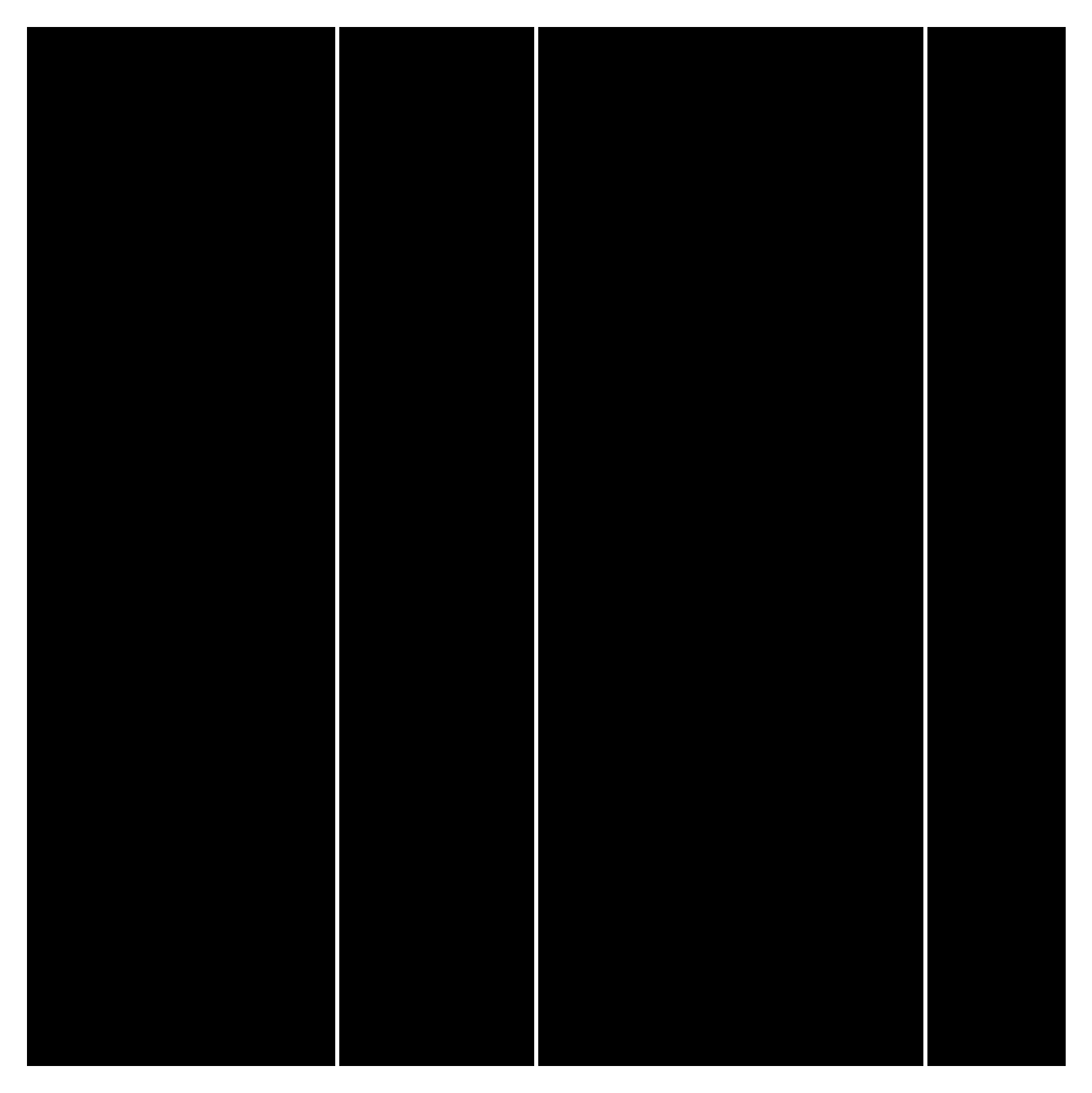} \\
(a) & (b) & (c) & (d) & (e) & (f) 
\end{tabular}
\caption{SAGE training illustration showing: (a) ground truth completely unseen velocity model, (b) corresponding migrated (RTM) image, (c) masked velocity observations (well-logs) with randomly observed wells, (d) mask indicating observed locations, (e) subsampled observed wells (f) subsampling mask indicating observed locations.}
\label{fig-sage-training-algorithm}
\end{figure*}

\begin{algorithm}[!t]
\caption{SAGE training}
\label{alg-sage-training}
\begin{algorithmic}
\State \textbf{Input:} $D_{\boldsymbol{\theta}},\, p(\mathbf{x}_{\mathrm{obs}},\mathbf{A},\mathbf{y}),\, p(\widetilde{\mathbf{A}}\mid\mathbf{A})$
\Repeat
  \State Sample $(\mathbf{x}_{\mathrm{obs}}, \mathbf{A}, \mathbf{y}) \sim p(\mathbf{x}_{\mathrm{obs}},\mathbf{A},\mathbf{y}),\;\widetilde{\mathbf{A}} \sim p(\widetilde{\mathbf{A}}\mid\mathbf{A})$
  \State Set $\widetilde{\mathbf{x}}_{\mathrm{obs}} \gets \widetilde{\mathbf{A}}\mathbf{x}_{\mathrm{obs}}$
  \State Sample $\sigma \sim \operatorname{LogNormal},\;\mathbf{n} \sim \mathcal{N}(\mathbf{0},\mathbf{I})$
  \State Set $\widehat{\mathbf{x}} \gets D_{\boldsymbol{\theta}}(\widetilde{\mathbf{x}}_{\mathrm{obs}} + \sigma\mathbf{n}, \mathbf{y}, \widetilde{\mathbf{A}}, \sigma)$
  \State Update $\boldsymbol{\theta} \gets \boldsymbol{\theta} - \eta \nabla_{\boldsymbol{\theta}} \|\mathbf{A}\widehat{\mathbf{x}} - \mathbf{x}_{\mathrm{obs}}\|_2^2$
\Until{convergence}
\end{algorithmic}
\end{algorithm}

We consider a setting in which direct access to prior velocity samples
\(\mathbf{x} \sim p(\mathbf{x})\), with \(x \in \mathbb{R}^N\) where
\(N=n_x \times n_z\) is unavailable. Instead, we observe only sparse,
but direct, measurements of the velocity field obtained from well logs
denoted by \(\mathbf{x}_\text{obs}\) together with corresponding seismic
summary statistics \(\mathbf{y}\) derived from RTM. Well-log
measurements correspond to column-wise observations of the velocity
field and can be represented as a masking operation. Let
\(\mathbf{w} \in \{0,1\}^{n_{x}}\) vector specifying which lateral
positions are observed. This induces a masking operator
\(\mathbf{A} \in \mathbb{R}^{N \times N}\) acting on unseen
\(\mathbf{x}\) defined as
\(A = \{\text{diag}(\mathbf{w} \otimes \mathbf{1}_{n_{z}}): \mathbf{w} \in \{0,1\}^{n_{x}}\}\),
where \(\otimes\) denotes the Kronecker product,
\(\mathbf{1}_{n_z} \in \mathbb{R}^{n_z}\) is a vector of ones and the
operator \(\mathrm{diag}(\cdot)\) maps a vector to a diagonal matrix
with that vector on its diagonal. The observed velocity measurements are
then given by \(\mathbf{x}_\text{obs} = \mathbf{A}\mathbf{x}\). In
practice, well locations vary across surveys; we model this variability
by treating the indicator vector as a random variable
\(\mathbf{w} \sim p(\mathbf{w})\), which in turn induces a stochastic
\(\mathbf{A} \sim p(\mathbf{A})\). Since only incomplete observations of
\(\mathbf{x}\) are available, direct characterization of the posterior
distribution \(p(\mathbf{x} \mid \mathbf{y})\) is intractable. This
limitation motivates the introduction of a proxy posterior distribution
\(\tilde{p}(\mathbf{x} \mid \mathbf{y})\) that can be learned from
jointly available RTM images and partial well-log measurements. Taking
note from the previous section, posterior estimation can be recast
within the denoising formulation of score-based networks. A naïve
approach is to enforce consistency of the denoiser with observed
velocity measurements through the measurement operator. This leads to
the following objective: \[
\hat{\boldsymbol{\theta}} 
= 
\arg\min_{\boldsymbol{\theta}} 
\mathbb{E}_{(\mathbf{y}, \mathbf{x}_{\mathrm{obs}}, \sigma, \mathbf{n}, \mathbf{A})}
\left\|
\mathbf{A}\, D_{\boldsymbol{\theta}}(\mathbf{x}_{\mathrm{obs}} + \sigma \mathbf{n}, \mathbf{y}, \mathbf{A}, \sigma)
- \mathbf{x}_{\mathrm{obs}}
\right\|_2^2,
\] where \(\mathbf{n} \sim \mathcal{N}(0, \mathbf{I})\) and
\(\mathbf{A} \sim p(\mathbf{A})\). However, this formulation admits a
trivial solution given by
\(D_{\boldsymbol{\theta}}(\cdot) = \mathbf{A}^\dagger \mathbf{x}_{\mathrm{obs}}\),
where \(\mathbf{A}^\dagger\) denotes pseudo-inverse of \(\mathbf{A}\).
This solution is undesirable, as it merely reconstructs the velocity
field at observed locations while leaving unobserved regions
unconstrained and effectively disregarding the conditioning information
provided by the RTM image \(\mathbf{y}\). To prevent convergence to the
trivial solution, we introduce a secondary masking (subsampling)
operator that acts as a stochastic submask of the original masking
operator. We define subsampling mask {[}24{]}: \[
\widetilde{\mathbf{A}} = \{\text{diag}(\tilde{\mathbf{w}} \otimes \mathbf{1}_{n_{z}}): \tilde{\mathbf{w}} \in \{0,1\}^{n_{x}}\} \ \text{with} \ \tilde{\mathbf{w}_i} \leq \mathbf{w}_i \ \forall \text{i = 1}, \ldots, n_{x}
\] where \(\tilde{\mathbf{w}}\) is a binary vector that selects a subset
of the originally observed columns. By construction,
\(\widetilde{\mathbf{A}}\) cannot reveal previously unobserved entries
and therefore constitutes a random submask of \(\mathbf{A}\). Applying
this operator yields a further subsampled observation as
\(\tilde{\mathbf{x}}_\text{obs} = \widetilde{\mathbf{A}}\mathbf{x}_\text{obs} = \widetilde{\mathbf{A}} \mathbf{A}\mathbf{x} = \widetilde{\mathbf{A}}\mathbf{x}\)
where the last equality follows from the diagonal structure of the
masking operators. We model this process as
\(\widetilde{\mathbf{A}} \sim p(\widetilde{\mathbf{A}}| \mathbf{A})\)
and
\(\tilde{\mathbf{x}}_\text{obs} = \widetilde{\mathbf{A}}\mathbf{x}\).
Using these subsampled observations, we define a training objective that
enforces global reconstruction while maintaining consistency with the
original measurements: \[
\hat{\boldsymbol{\theta}}
= \arg\min_{\boldsymbol{\theta}}
\mathbb{E}_{p_{\mathrm{train}}}
\left\|
\mathbf{A}D_{\boldsymbol{\theta}}(\tilde{\mathbf{x}}_{\mathrm{obs}}+\sigma\mathbf{n},\mathbf{y},\widetilde{\mathbf{A}},\sigma)
-\mathbf{x}_{\mathrm{obs}}
\right\|_2^2
\] with
\(p_{\mathrm{train}}=p(\mathbf{y},\mathbf{x}_{\mathrm{obs}},\tilde{\mathbf{x}}_{\mathrm{obs}},\sigma,\mathbf{n},\mathbf{A},\widetilde{\mathbf{A}})\).
The training procedure is summarized in {Algorithm
Algorithm~\ref{alg-sage-training}}, while the matrix representations of
the observations and the diagonal of masking operators are illustrated
in {Figure \ref{fig-sage-training-algorithm}}.A key characteristic of
SAGE is that it simultaneously performs generative modeling and
inpainting, enabling the reconstruction of globally consistent velocity
fields from partial observations. Once training is complete, posterior
sampling is performed by initializing the unobserved components with
noise and iteratively applying the learned denoising dynamics
conditioned on the RTM image. This procedure yields samples from the
learned proxy posterior distribution.

\section{Result and Numerical Study}\label{result-and-numerical-study}

\subsection{SAGE Inference on synthetic
wells}\label{sage-inference-on-synthetic-wells}

\begin{figure}[!t]
\centering
\begin{tabular}{@{}cc@{}}
\includegraphics[width=0.48\columnwidth]{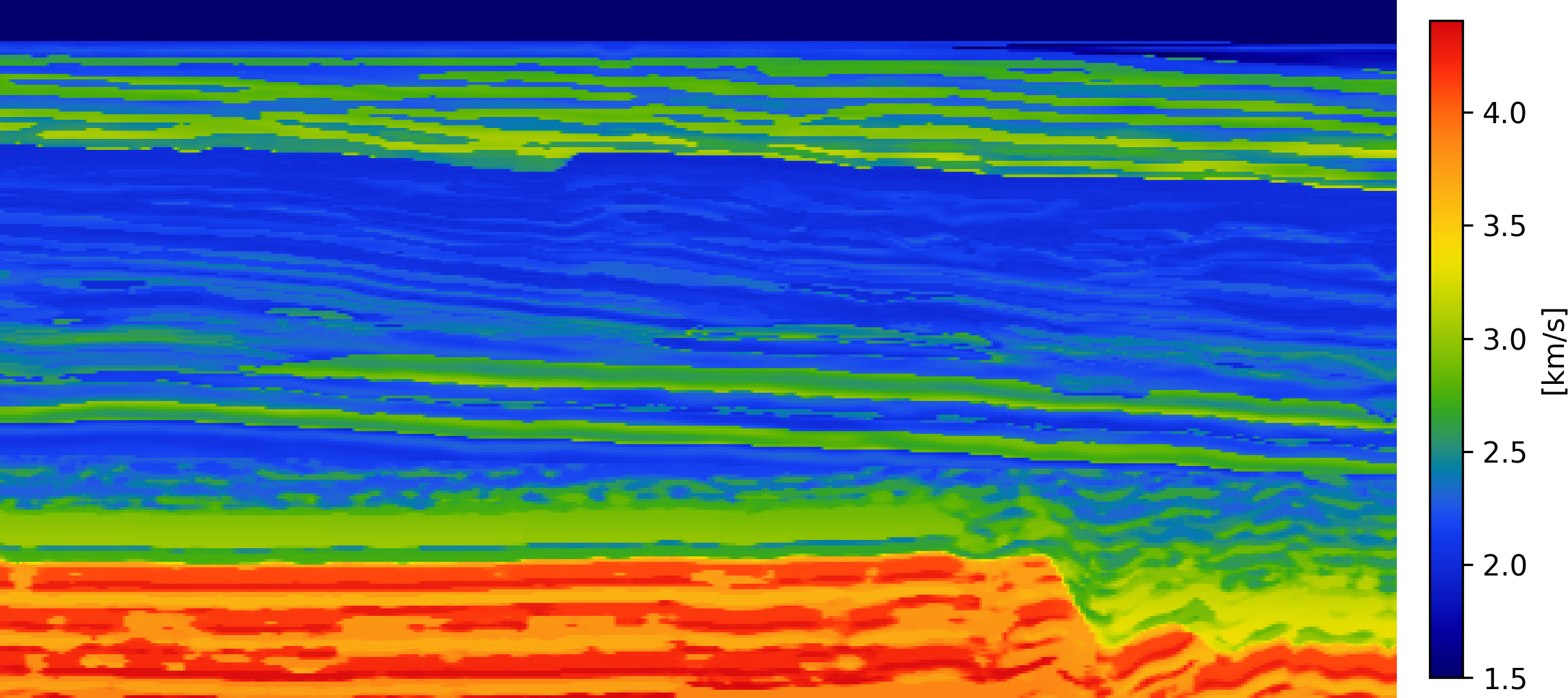} &
\includegraphics[width=0.48\columnwidth]{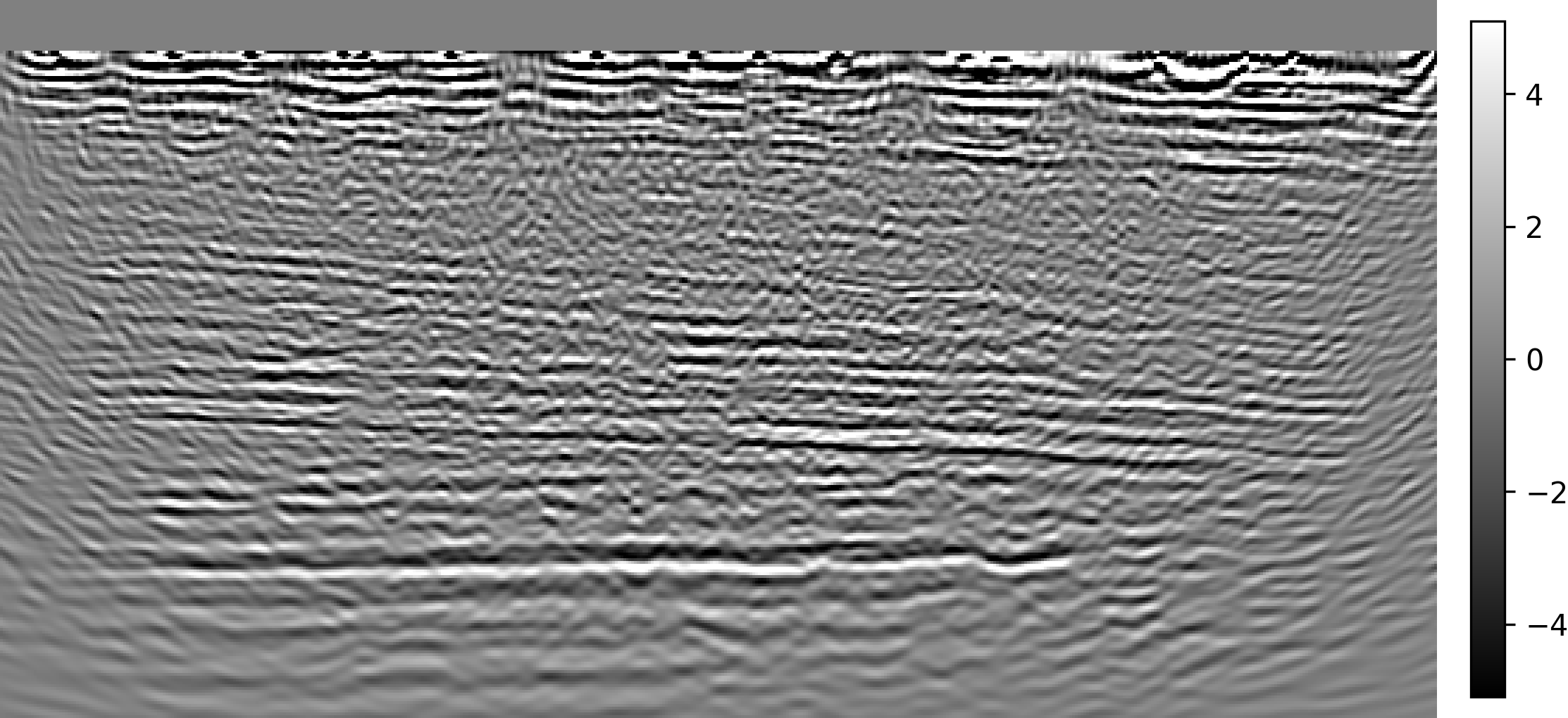} \\
(a) Ground truth velocity & (b) RTM image \\[1ex]
\includegraphics[width=0.48\columnwidth]{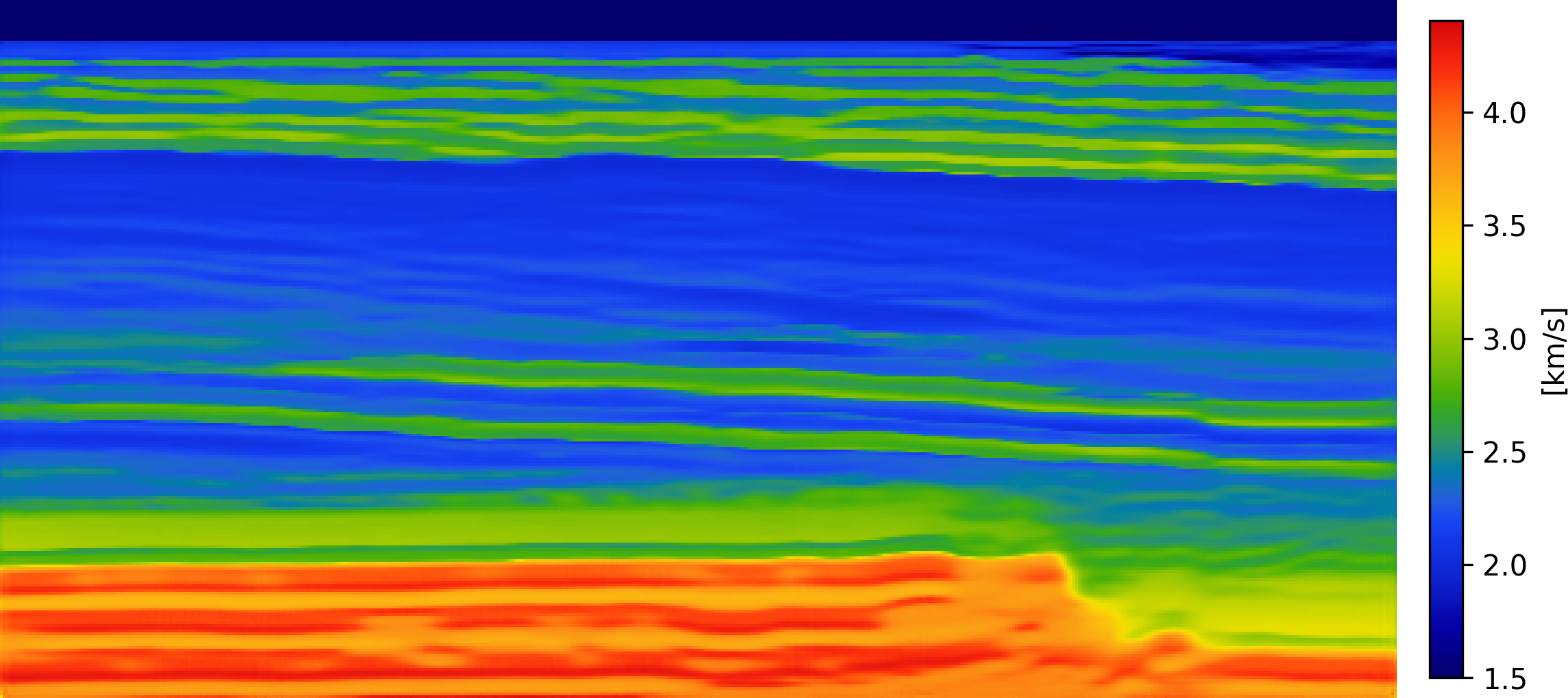} &
\includegraphics[width=0.48\columnwidth]{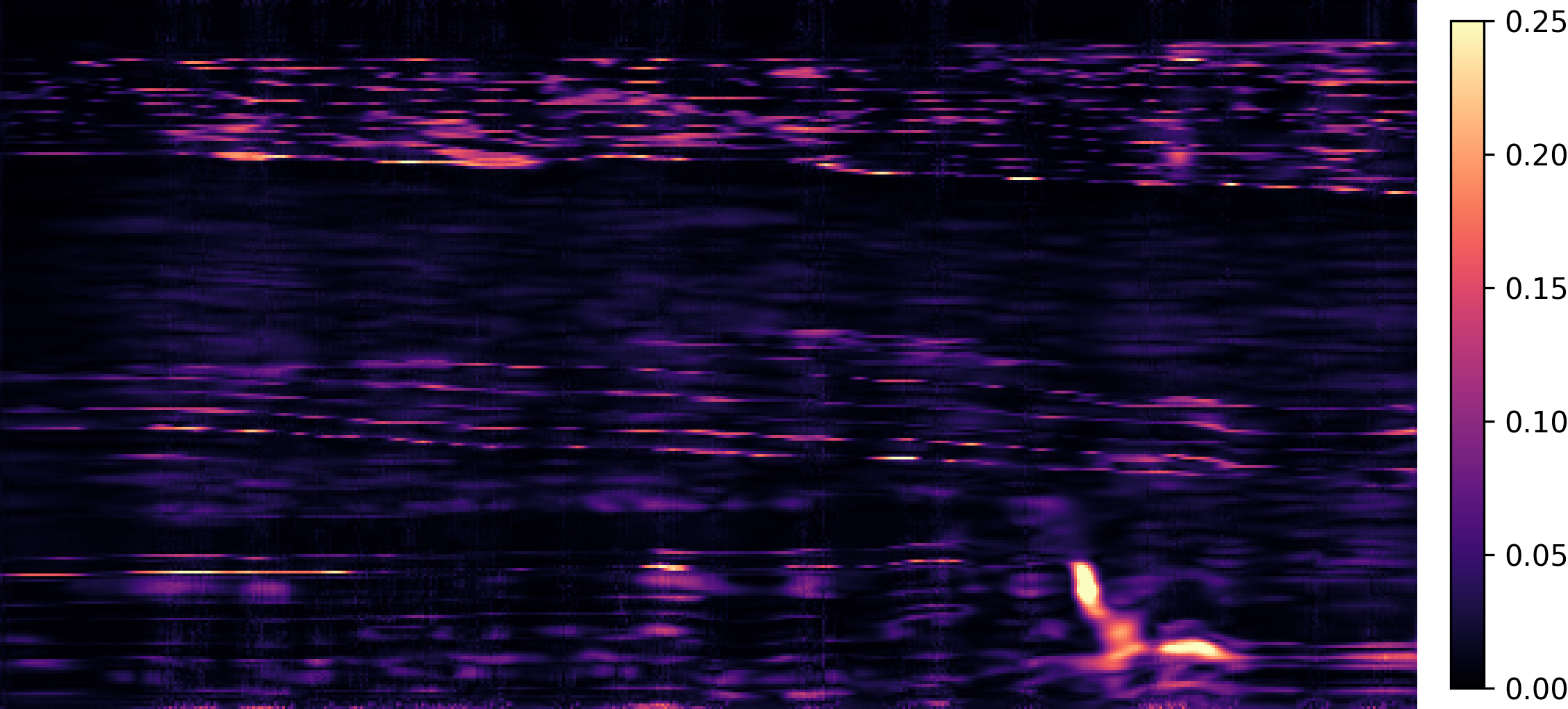} \\
(c) SAGE Posterior mean & (d) SAGE Posterior std \\
\end{tabular}
\caption{SAGE results showing: (a) ground truth velocity model, (b) corresponding RTM image used as conditioning, (c) posterior mean, and (d) posterior std over 16 samples.}
\label{fig-sage-results}
\end{figure}

To evaluate SAGE, first we slice synthetic 2D Earth acoustic velocity
models derived from the 3D Compass model, which is representative of
geological formations in the North Sea region {[}25{]}. The dataset
consists of 1000 velocity realizations, each defined on a
\(256\times 512\) grid with a spatial resolution of \(12.5\) meters,
covering an area of \(3.2km\times 6.4\mathrm{km}\). Then for each
velocity model, seismic data are simulated using \(16\) sources and
\(256\) receivers, with Ricker wavelet centered at the dominant
frequency of \(20\mathrm{Hz}\) and a recording duration of \(3.2\)
seconds. To simulate realistic acquisition conditions, \(10\mathrm{dB}\)
colored Gaussian noise is added to the shot records. RTM images are
generated using a Gaussian-smoothed 2D background velocity model. Both
wave simulation and migration are implemented using JUDI {[}26{]}. This
procedure yields a dataset of paired RTM images and corresponding
full-resolution velocity models. However, in the SAGE setting, access to
complete velocity models is not assumed. Therefore, we retain 5 out of
256 columns for each model and zero out the remaining entries,
effectively removing approximately 99\% of the information. These
retained columns represent sparsely distributed well-log measurements
that are spatially co-located (tied) with the RTM images.

SAGE is trained using the procedure described in {Algorithm
Algorithm~\ref{alg-sage-training}}. During training, the RTM image and
corrupted subsampling mask are concatenated with the noisy, partially
observed velocity field and provided as input to the network. A U-Net
architecture is employed as the denoiser, and the total training time is
approximately 20 GPU hours. For evaluation, we select unseen RTM
examples withheld during training and generate samples from the learned
proxy posterior using the trained network. {Figure
\ref{fig-sage-results}} presents the inference results obtained with
SAGE. The posterior mean accurately reconstructs the dominant features
of the ground-truth velocity model, achieving a structurally consistent
estimate with an SSIM score of 0.82. The posterior standard deviation
further provides a meaningful quantification of uncertainty, with
elevated variance localized in regions characterized by increased
geological complexity. We emphasize that, although comparisons are made
against the ground truth for evaluation purposes, SAGE is trained
exclusively on partially observed velocity fields and has no access to
complete models during training. Consequently, mild smoothing of
fine-scale structures is expected and reflects the inherent ambiguity
induced by incomplete observations rather than a deficiency of the
network.

\subsection{Training of WISE with SAGE
samples}\label{training-of-wise-with-sage-samples}

\begin{figure*}[!t]
\centering
\includegraphics[width=\textwidth]{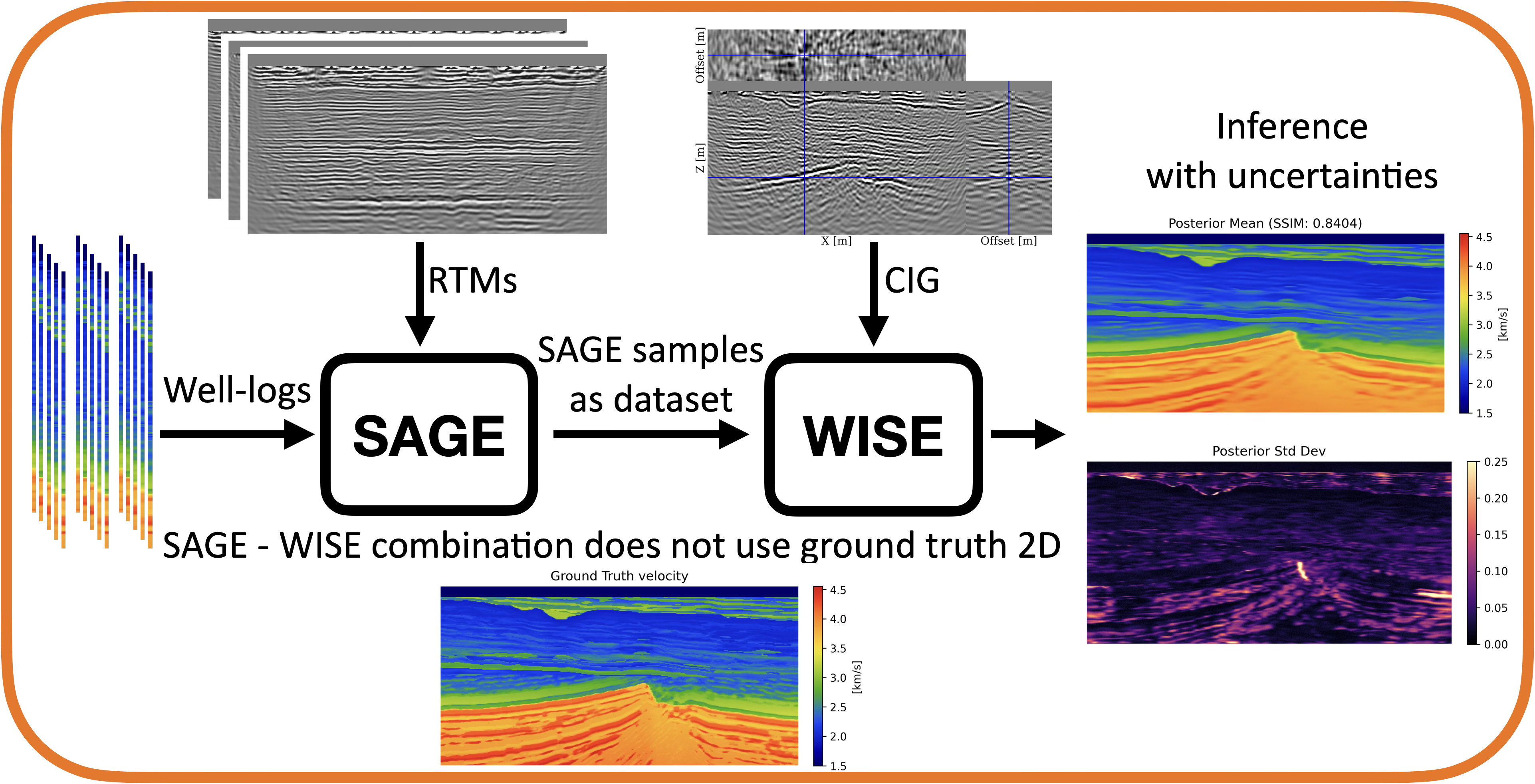}
\caption{Workflow of training WISE with SAGE samples.}
\label{fig-workflow}
\end{figure*}

A prominent downstream application of SAGE is the generation of
synthetic velocity samples for training specialized inversion networks.
In this work, we consider WISE {[}11{]}, a variational inference-based
framework for FWI that produces ensembles of migration velocity models
conditioned on common-image gathers (CIGs). The key advantage of WISE
lies in its use of high-dimensional CIGs during inference, which provide
stronger kinematic constraints than RTM images and are closer to a
unitary mapping even under imperfect migration velocities {[}27{]}.
Despite this advantage, WISE requires access to large collections of
high-quality, fully observed velocity models during training, which
limits its applicability in data-scarce settings. To address this
limitation, we generate velocity samples from SAGE for unseen test
examples and use these samples as a training dataset. The overall
workflow is illustrated in {Figure \ref{fig-workflow}}. Inference
results indicate that WISE trained on SAGE samples yields high-quality
velocity estimates, as reflected in both posterior mean and associated
uncertainty maps. Notably, compared to training on ground-truth velocity
models, the use of SAGE-generated samples incurs only a modest
degradation in performance, with SSIM decreasing from 0.88 to 0.84.

\subsection{SAGE inference on real seismic images and well
data}\label{sage-inference-on-real-seismic-images-and-well-data}

\begin{figure*}[!t]
\centering
\includegraphics[width=\textwidth]{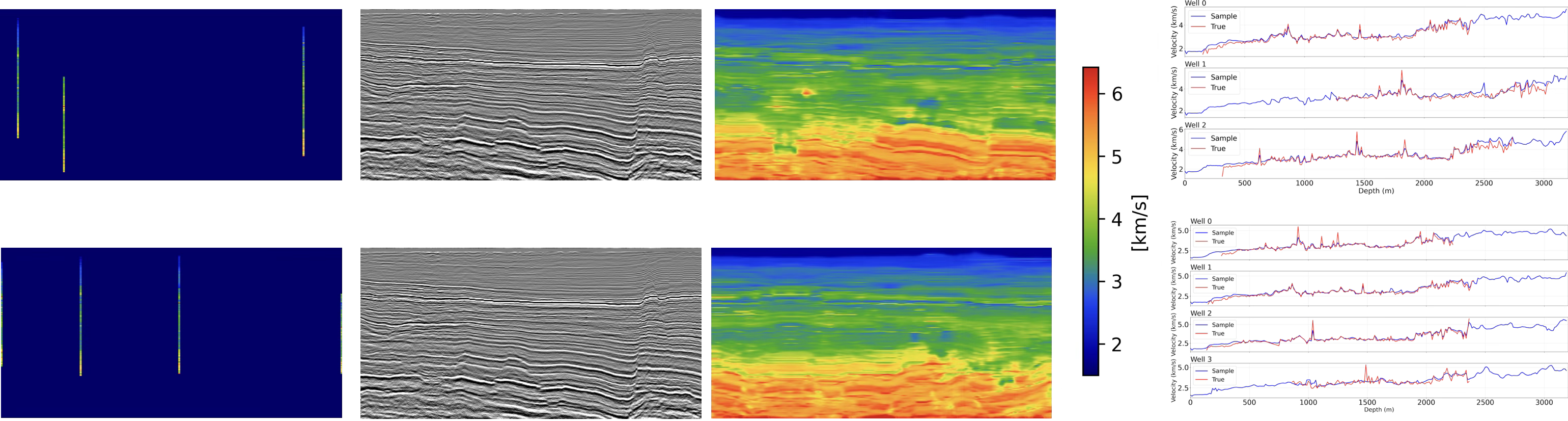}
\caption{SAGE inference on real migrated seismic image: (a) ground truth well-logs, (b) migrated seismic images, (c) velocity samples constructed by SAGE, (d) trace comparison of SAGE samples vs. ground truth well-logs.}
\label{fig-real-data}
\end{figure*}

We evaluate the performance of SAGE on real seismic and well-log
obtained from the UK National Data Repository {[}28{]}. Following an
preprocessing pipeline, including checkshot based time-to-depth
conversion, well tie, quasi-2D line extraction, and kinematically
consistent downsampling, we curate a dataset for training and
evaluation. Additional details on the data preparation can be found in
{[}29{]}. Due to the limited availability of well-log measurements (40
wells), we adopt a fine-tuning strategy in which a network pre-trained
on the synthetic Compass model is adapted to the real data. {Figure
\ref{fig-real-data}} presents the preliminary inference results of SAGE
on real seismic data. The generated velocity models exhibit complex
geological structure, aligning with the features observed in the
migrated images while remaining consistent with independent well-log
measurements not used during inference. We anticipate further
improvements in performance as the training dataset is expanded.

\section{Conclusion and Future Work}\label{conclusion-and-future-work}

We introduced SAGE, a framework for subsurface velocity generation from
incomplete observations, leveraging well-logs and migrated images. SAGE
learns a proxy posterior during training and, at inference, produces
full velocity fields conditioned solely on seismic migrated images, with
well information implicitly encoded. Experiments demonstrate that SAGE
produces high-quality and geologically consistent velocity models that
can be utilized for downstream tasks, including training specialized
learned inversion frameworks such as WISE. Notably, even with severely
limited well-log data, fine-tuning on real seismic observations yields
statistically plausible velocity realizations. Future work will extend
SAGE to three-dimensional settings and larger curated field datasets.

\section{Acknowledgement}\label{acknowledgement}

This research was carried out with the support of Georgia Research
Alliance, partners of the ML4Seismic Center. We acknowledge that we used
data from UK National Data Repository and it contains information
provided by the North Sea Transition Authority and/or other third
parties. The authors used ChatGPT to refine sentence structures and
improve readability. After using this service, the authors reviewed and
edited the content as needed and take full responsibility.

\section{References}\label{references}

\phantomsection\label{refs}

\end{document}